%% file: main.tex
\documentclass[10pt,twocolumn,letterpaper]{article}

\usepackage{cvpr}              

\input{preamble}

\definecolor{cvprblue}{rgb}{0.21,0.49,0.74}
\usepackage[pagebackref,breaklinks,colorlinks,citecolor=cvprblue]{hyperref}
\usepackage{xspace}
\usepackage{booktabs}
\usepackage{multirow}
\usepackage{rotating}
\usepackage{color}
\usepackage[title]{appendix}
\usepackage{makecell}
\usepackage{pifont}
\usepackage{float}

\def\paperID{2038} 
\def\confName{CVPR}
\def\confYear{2024}

\title{ParameterNet: Parameters Are All You Need}

\author{Kai Han$^{1,*}$ \hspace{0.23cm} Yunhe Wang$^{1,*}$ \hspace{0.23cm} Jianyuan Guo$^{1,2,}$\thanks{Equal contribution.} \hspace{0.23cm} Enhua Wu$^{3,4}$\\
$^1$Huawei Noah's Ark Lab \hspace{0.23cm} $^2$The University of Sydney\\
$^3$State Key Lab of Computer Science, ISCAS \hspace{0.23cm} $^4$University of Macau\\
{\tt\small \{kai.han,yunhe.wang,jianyuan.guo\}@huawei.com, weh@ios.ac.cn}
}

\begin{document}
\maketitle

\begin{abstract}
The large-scale visual pretraining has significantly improve the performance of large vision models. However, we observe the \emph{low FLOPs pitfall} that the existing low-FLOPs models cannot benefit from large-scale pretraining.
In this paper, we introduce a novel design principle, termed ParameterNet, aimed at augmenting the number of parameters in large-scale visual pretraining models while minimizing the increase in FLOPs. We leverage dynamic convolutions to incorporate additional parameters into the networks with only a marginal rise in FLOPs. The ParameterNet approach allows low-FLOPs networks to take advantage of large-scale visual pretraining. Furthermore, we extend the ParameterNet concept to the language domain to enhance inference results while preserving inference speed.
Experiments on the large-scale ImageNet-22K have shown the superiority of our ParameterNet scheme. For example, ParameterNet-600M can achieve higher accuracy on ImageNet than the widely-used Swin Transformer (81.6\% \emph{vs.} 80.9\%) and has much lower FLOPs (0.6G \emph{vs.} 4.5G). In the language domain, LLaMA-1B enhanced with ParameterNet achieves 2\% higher accuracy over vanilla LLaMA. The code will be released at \url{https://parameternet.github.io/}.
\end{abstract}

\section{Introduction}
\label{sec:intro}
Thanks to advancements in computational hardware and data engineering, large-scale visual pretraining has witnessed remarkable progress as a fundamental component in computer vision. Pretrained vision models act as efficient representation learners, showcasing their utility in various downstream visual tasks, including image recognition~\cite{imagenet,efficientnetv2}, object detection~\cite{swin,pvt} and semantic segmentation~\cite{he2017mask,setr}.

The mainstream pretrained vision models usually requires a large amount of resources including data, parameters and FLOPs. These three key factors heavily influence the performance and basically follow the scaling law~\cite{zhai2022scaling}. The large pretraining data can provide diverse samples for representation learning. The sizes of these datasets range from millions~\cite{imagenet,kuznetsova2020open} to billions~\cite{sun2017revisiting,thomee2015new}, for example, the widely-used ImageNet-22K dataset~\cite{imagenet} consists of 14M images and 21,841 categories. To better fitting on the large dataset, the model sizes (including both parameters and FLOPs) are getting larger and larger in recent years, \eg, ViT-G/14 model has 1.8B parameters and 965B FLOPs~\cite{zhai2022scaling}.

\begin{figure*}[htp]
	\centering
	\includegraphics[width=0.5\linewidth]{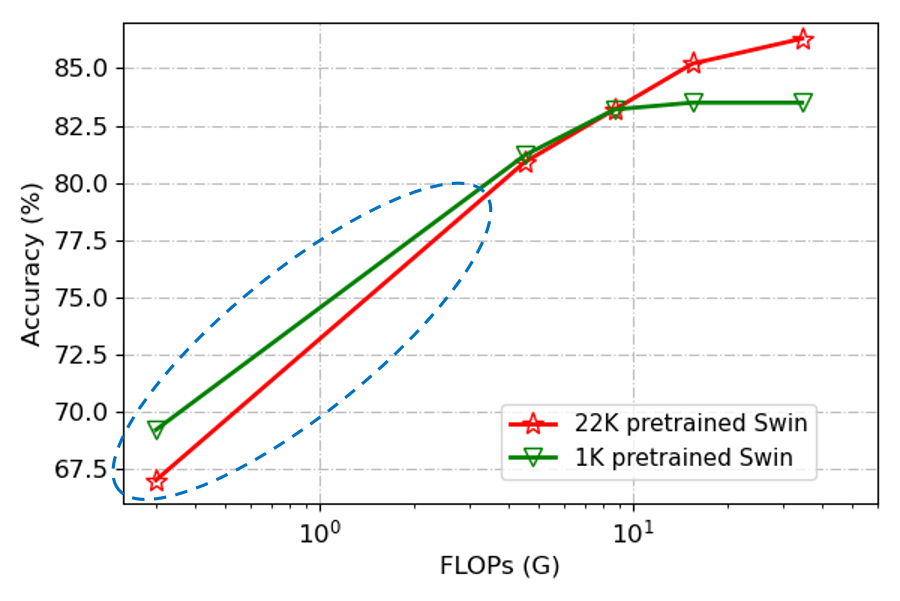}
	\renewcommand\arraystretch{1.05}
	\resizebox{.9\columnwidth}{!}{
		\begin{tabular}[b]{l|cccc}
			\toprule[1.5pt]
			Model & Pretrain data & \#Params & \#FLOPs & Top1 \\
			\midrule
			Swin-300M & IN-1K & 1.9M & 0.3G & 69.2 \\
			Swin-T & IN-1K & 28M & 4.5G & 81.2 \\
			Swin-S & IN-1K & 50M & 8.7G & 83.2 \\
			Swin-B & IN-1K & 88M & 15.4G & 83.5 \\
			Swin-L & IN-1K & 197M & 34.5G & 83.5 \\
			\midrule
			Swin-300M & IN-22K & 1.9M & 0.3G & 67.0 \\
			Swin-T & IN-22K & 28M & 4.5G & 80.9 \\
			Swin-S & IN-22K & 50M & 8.7G & 83.2 \\
			Swin-B & IN-22K & 88M & 15.4G & 85.2 \\
			Swin-L & IN-22K & 197M & 34.5G & 86.3 \\
			\bottomrule[1pt]
			\multicolumn{5}{c}{\vspace{1em}}\\
	\end{tabular}}
	\caption{Low FLOPs pitfall. Swin Transformer results on ImageNet-1K validation set. The red and blue lines denote ImageNet-22K and ImageNet-1K pretraining, respectively.}
	\label{fig:swin}
\end{figure*}

\begin{figure*}[htp]
	\centering
	\includegraphics[width=0.5\linewidth]{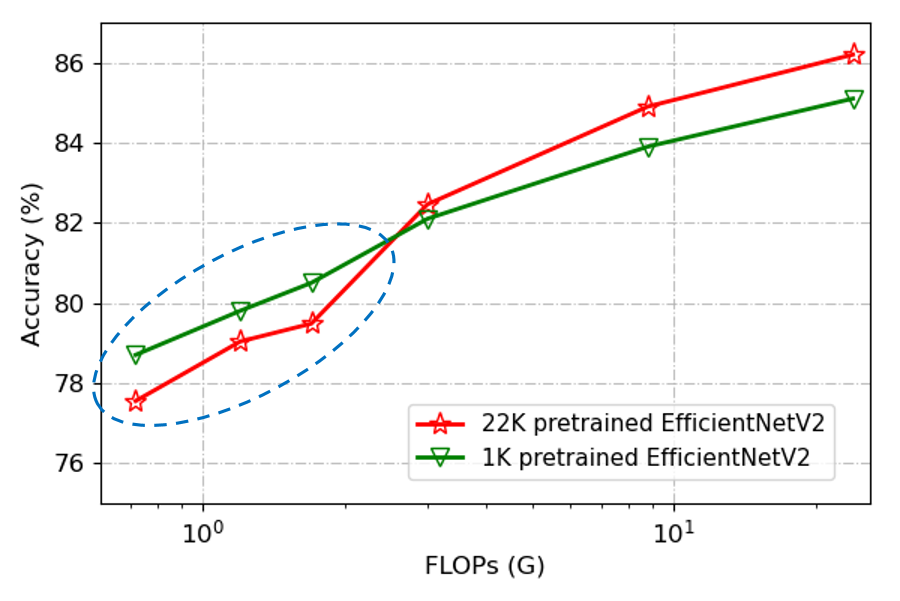}
	\renewcommand\arraystretch{1.05}
	\resizebox{.9\columnwidth}{!}{
		\begin{tabular}[b]{l|cccc}
			\toprule[1.5pt]
			Model & Pretrain data & \#Params & \#FLOPs & Top1 \\
			\midrule
			EfficientNetV2-B0 & IN-1K & 7.1M & 0.72G & 78.7 \\
			EfficientNetV2-B1 & IN-1K & 8.1M & 1.2G & 79.8 \\
			EfficientNetV2-B2 & IN-1K & 10.1M & 1.7G & 80.5 \\
			EfficientNetV2-B3 & IN-1K & 14.4M & 3.0G & 82.1 \\
			EfficientNetV2-S & IN-1K & 21.5M & 8.4G & 83.9 \\
			EfficientNetV2-M & IN-1K & 54.1M & 24.7G & 85.2 \\
			\midrule
			EfficientNetV2-B0 & IN-22K & 7.1M & 0.72G & 77.6 \\
			EfficientNetV2-B1 & IN-22K & 8.1M & 1.2G & 79.0 \\
			EfficientNetV2-B2 & IN-22K & 10.1M & 1.7G & 79.5 \\
			EfficientNetV2-B3 & IN-22K & 14.4M & 3.0G & 82.5 \\
			EfficientNetV2-S & IN-22K & 21.5M & 8.4G & 84.9 \\
			EfficientNetV2-M & IN-22K & 54.1M & 24.7G & 86.2 \\
			\bottomrule[1pt]
			\multicolumn{5}{c}{\vspace{1.5em}}\\
	\end{tabular}}
	\caption{Low FLOPs pitfall. EfficientNetV2 results on ImageNet-1K validation set. The red and blue lines denote ImageNet-22K and ImageNet-1K pretraining, respectively.}
	\label{fig:enetv2}
\end{figure*}

The visual applications on mobile devices usually requires fast inference, so it is difficult to deploy the existing pretrained vision models due to the high computational cost. To address this issue, we empirically study the effect of FLOPs in large-scale visual pretraining. ImageNet-22K is adopted as the large-scale pretraining data and ImageNet-1K is a relatively small dataset for comparison. 
The pretrained transformer and CNN models are then finetuned on ImageNet-1K to evaluate the performance. As shown in Figure~\ref{fig:swin} and \ref{fig:enetv2}, when model FLOPs gradually increase, the model accuracy increases consistently. For the high-FLOPs models, 22K pretrained models outperform 1K ones. However, the low-FLOPs models cannot benifit from large-scale pretraining, and we called this observation as \emph{low FLOPs pitfall}. 

In this paper, we construct low-FLOPs ParameterNet by adding more parameters while maintaining low FLOPs for large-scale visual pretraining. It is a general design principle and there are various approaches with more parameters and low FLOPs. For instance, here we mainly consider the efficient dynamic convolution which manyfold increase the number of parameters while almost does not bring in extra FLOPs. The ParameterNet scheme can enable the previous networks to benefit from the large-scale visual pretraining and overcome the \emph{low FLOPs pitfall}. In the experiments, the ImageNet-22K pretrained ParameterNets can improve the performance by about +2\% over the regular ImageNet-1K training. For example, ParameterNet-600M achieves 81.6\% top-1 accuracy on ImageNet-1K val set whose \#FLOPs is 7$\times$ lower than that of Swin-T. The proposed ParameterNet can be extended to the large language model (LLM) domain and experiments on LLaMA model~\cite{llama} verify the effectiveness.

\begin{figure*}[htp]
	\centering
	\includegraphics[width=0.9\linewidth]{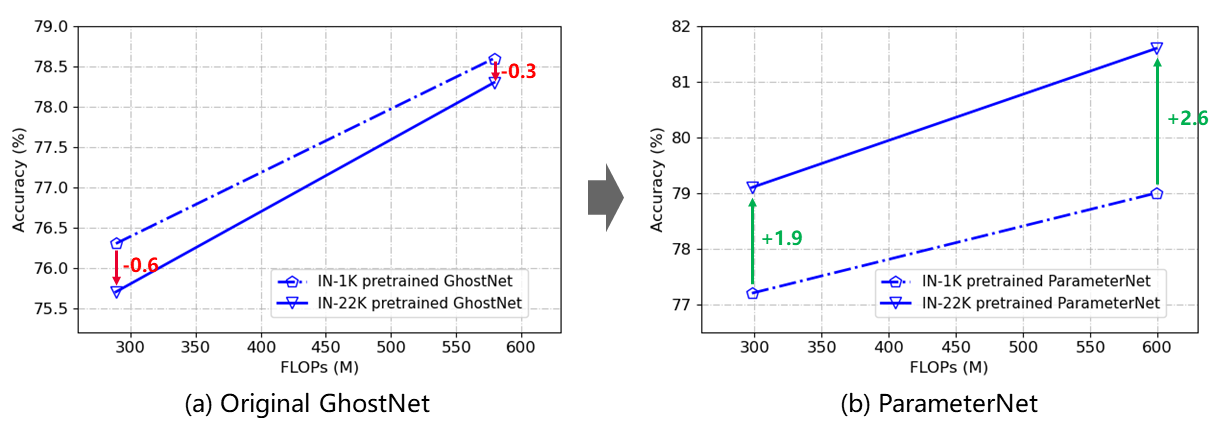}
	\caption{Results on ImageNet-1K validation set. The original GhostNet falls into the \emph{low FLOPs pitfall}. The proposed ParameterNet overcomes the \emph{low FLOPs pitfall}.}
	\label{fig:main}
\end{figure*}

The main contributions of this paper can be summarized as follows.
\begin{itemize}
	\item We observe an interesting phenomenon in large-scale visual pretraining called the \emph{low FLOPs pitfall}, that is, the performances of high-FLOPs models increase with more training data, but the models with low-FLOPs.
	\item We propose that parameters are more important than FLOPs for large-scale visual pretraining and further introduce the ParameterNet scheme by adding more parameters while maintaining low FLOPs.
	\item The proposed ParameterNet scheme can overcome the \emph{low FLOPs pitfall}, and experimental results on vision and language tasks show that ParameterNet achieves significantly higher performance with large-scale pretraining.
\end{itemize}

\section{Related Work}
In this section, we briefly revisit the related works about visual backbone networks and visual pretraining.

\paragraph{Visual Backbone Networks.}
The deep neural networks in computer vision can be divided into CNNs, vision transformers and others. CNN used to be the mainstream network architecture for visual tasks~\cite{lecun1998gradient,alexnet,resnet}. The first trainable CNN \ie, LeNet~\cite{lecun1998gradient} is applied on optimal character recognition (a typical visual task). From 2012, CNNs began to be deeper and larger for more complex visual tasks, such as image classification~\cite{alexnet}, object detection~\cite{fasterRCNN} and semantic segmentation~\cite{fcn}. ResNet~\cite{resnet} introduces the shortcut connection to train the deeper networks and is widely used in vision and other communities. MobileNet~\cite{mobilenet} is designed for mobile devices and EfficientNet~\cite{efficientnet} scales the network from small to large.

Vision transformer is introduced into visual tasks from 2020~\cite{vit-survey,vit,detr}. ViT~\cite{vit} is the first transformer backbone by dividing the image into patches and processing them using the standard transformer architcture. Then a number of variants and improvements are proposed incluidng the pyramid architectures~\cite{pvt,swin}, the local attentions~\cite{tnt,swin} and hybrid networks~\cite{guo2022cmt,metaformer}.

Beyond CNNs and transformers, other types of neural networks are also explored for visual tasks. MLP-like architectures~\cite{mixer,resmlp} with only fully-connected layers as main operators can potentially simplify the software and hardware design for AI. The improved versions of MLP~\cite{cyclemlp,asmlp,hire,wavemlp} can enhance locality and translation equivalence. GNN has also been expored in vision and achieves competitive performance to transformers~\cite{vig}. The pretrained backbone neteworks help much for the downstream visual tasks, and the study on the pretraining of backbones is an important topic.

\paragraph{Visual Pretraining.}
Large-scale pretraining has achieved great success on natural language processing such as GPT series~\cite{gpt3,gpt4}. In the field of computer vision, large-scale pretraining is also beneficial and helps for downstream tasks~\cite{bit,swin,sam}.
The large datasets are the foundations of pretraining. To distinguish from the regular ImageNet-1K training, we consider the dataset with an order of magnitude more than ImageNet-1K (\ie, more than 10M samples) as the large-scale dataset. The commonly-used large visual datasets include ImageNet-22K~\cite{imagenet}, JFT-300M~\cite{sun2017revisiting}, YFFCC100M~\cite{yfcc100m} and IG-1B-Targeted~\cite{yalniz2019billion}.
The supervised pretraining is popular as it can learn semantic and meaningful representations for downstream visual tasks like segmentation and detection. BiT~\cite{bit} pretrained on JFT-300M achieves state-of-the-art on fine-grained recognition datasets. The ImageNet-22K pretrained Swin Transformer~\cite{swin} obtains high performance on segmentation and detection tasks.
The unsupervised pretraining appeals many researchers as it does not requires labels and may leverage the massive unlabeled data. The most popular approaches of visual unsupervised pretraining are contrastive learning~\cite{he2020momentum,simclr} and masked image modeling (MIM)~\cite{bao2021beit,he2022masked,xie2022simmim}.
In this paper, we utilize the vanilla supervised pretraining for simplicity and generality.

\section{Low FLOPs Pitfall}
The computational cost (\ie, FLOPs) is an important term in the scaling of visual models. We first investigate the effect of FLOPs and observe inspiring phenomenon. Both transformer and CNN architectures are studied on ImageNet-22K and ImageNet-1K pretraining.

\paragraph{Transformer.}
Swin Transformer~\cite{swin} is a representative vision transformer architecture with window attention and shifted window. We reproduce the models using the official code\footnote{https://github.com/microsoft/Swin-Transformer} and pretrain Swin Transformers with different scales on both ImageNet-22K and ImageNet-1K. The ImageNet-1k finetuning results are reported in Figure~\ref{fig:swin} for comparison. From the results, we can see that the accuracy increases as the FLOPs increase gradually with both ImageNet-1K and ImageNet-22K pretraining. For models with high FLOPs ($>$10G), pretraining on ImageNet-22K outperforms that on ImageNet-1K. However, pretraining on more data does not improve the performance for models with lower FLOPs ($<$4G).

\paragraph{CNN.}
For CNN, we select the widely-used EfficientNetV2~\cite{efficientnetv2} which is a family of convolutional networks scaling from small to large. We use the official code\footnote{https://github.com/google/automl/tree/master/efficientnetv2} and pretrain the models on both ImageNet-22K and ImageNet-1K. From the ImageNet-1k finetuning results in Figure~\ref{fig:enetv2}, we can observe the similar trend as that in Swin Transformer, especially, EfficientNetV2 models with less than 2G FLOPs pretraining on ImageNet-22K cannot perform better than those pretraining on ImageNet-1K.

From the observations of both transformer and CNN networks, we have a empirical conclusion that low-FLOPs models cannot benefit from large-scale pretraining, which is named as \emph{low FLOPs pitfall}.

\section{Approach}
In this section, we investigate the low-FLOPs networks under large-scale pretraining setting. 

\subsection{Architecture: Transformer vs. CNN}
Here we do not propose a new architecture and select the most suitable low-FLOPs network architecture for large-scale visual pretraining.
ViT~\cite{vit} and its variants~\cite{swin,tnt,pvt} have shown the superiority of transformer over CNN in the field of large vision models. As shown in the appendix, Transformer-based models consistently outperform CNNs with similar computational cost when the FLOPs are higher than 5G FLOPs. As for smaller models especially mobile-level model within 600M FLOPs, CNN with inductive bias including locality and s translation equivariance remain dominant. To build efficient backbones for visual tasks, we select CNN as the base model. GhostNet~\cite{ghostnet} is the representative state-of-art mobile model which introduces cheap operation to simplify the standard convolutional layer.

\subsection{Parameters Are All You Need}
The number of parameters and FLOPs are highly corelated in neural networks. The model with large number of parameters usually has high FLOPs. Considering the intuition that large data requires more parameters, we construct ParameterNet by adding parameters while maintaining low FLOPs.

We start from the conventional convolutional layer. Given the input feature $X\in\mathbb{R}^{C_{in}\times H\times W}$ and the weight tensor $W\in\mathbb{R}^{C_{out}\times C_{in}\times K\times K}$, the conventional convolutional layer operates as
\begin{equation}
	Y = X * W,
\end{equation}
where $Y\in\mathbb{R}^{C_{out}\times H'\times W'}$ is the output, $*$ is the convolution operation and the bias term is omitted for concision. The fully-connected layer can be viewed as the convolutional layer with $1\times1$ kernel size.

\begin{figure}[htb]
	\centering	
	\setlength{\tabcolsep}{2pt}{
		\begin{tabular}{cc}
			\makecell*[c]{\includegraphics[width=0.35\linewidth]{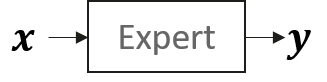}}  & \makecell*[c]{\includegraphics[width=0.65\linewidth]{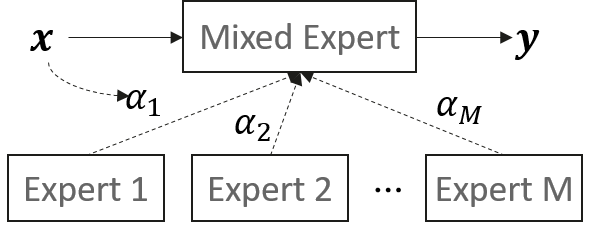}} 
			\\
			\small (a) Normal layer & \small (b) MoE layer
		\end{tabular}
	}
	\caption{The MoE layer can add more parameters while maintain low FLOPs.}
	\label{fig:moe}
\end{figure}

Our design principle is adding more parameters while maintaining low FLOPs. Thus, we introduce the parameter augmentation function which aims to introduce more parameters:
\begin{equation}
	W' = f(W).
\end{equation}
This function $f$ should satisfy two basic rules: 1) it does not require much computational cost, and 2) it can largely increase the model capacity or trainable parameters.
There are various approaches to construct ParameterNet, such as dynamic convolution~\cite{dynamic-conv} and re-parameterized convolution~\cite{repvgg}. Although the re-parameterized convolution increase the number of parameters during training, its parameters and FLOPs are unchanged for inference, that is, the model capacity is not increased. In this paper, we mainly consider the efficient dynamic convolution (a type of MoE layer in Figure~\ref{fig:moe}) which manyfold increase the number of parameters while almost does not bring in extra FLOPs. 

The dynamic convolution~\cite{dynamic-conv} with $M$ dynamic experts can be written as
\begin{equation}
	\begin{aligned}
		Y &= X * W',\\
		W' &= \sum_{i=1}^{M}\alpha_iW_i.
	\end{aligned}
\end{equation}
where $W_i\in\mathbb{R}^{C_{out}\times C_{in}\times H\times W}$ is the $i$-th convolutional weight tensor and $\alpha_i$ is the corresponding dynamic coefficient. The coefficient $\alpha_i$ is dynamically generated \wrt different input samples, and a typical manner is generating based on the input using MLP module. For the input $X$, a global average pooling is applied to fuse the information into a vector and then a two-layer MLP module with softmax activation is used to produce the coefficients dynamically:
\begin{equation}\label{eq:coefficient}
	\alpha = \textit{softmax}(\textit{MLP}(\textit{Pool}(X))),
\end{equation}
where $\alpha\in\mathbb{R}^{M}$. The coefficient generation in Eq.~\ref{eq:coefficient} only brings a nelegable FLOPs compared to the original convolutional layer. In this way, ParameterNet implemented with dynamic convolution can largely introducing much more parameters while minimizing the increase in FLOPs.

\begin{table}[htp]
	\vspace{-0em}
	\centering
	\small
	\caption{Training hyper-parameters on ImageNet datasets.}
	\label{table-hyper}
	\renewcommand\arraystretch{1.1}
	\setlength{\tabcolsep}{3pt}
	\begin{tabular}{l|ccc}
		\toprule[1.5pt]
		Config & ImageNet-1K & ImageNet-22K & Finetuning \\ 
		\midrule
		Epochs & 300 & 90 & 30 \\
		Optimizer & AdamW & AdamW & AdamW \\
		Batch size & 1024 & 4096 & 512 \\
		Start learning rate & {1e-3} & 4e-3 & 5e-4 \\
		Layer decay & \ding{55} & \ding{55} & 0.5 \\
		LR schedule & {Cosine} & Cosine & Cosine \\
		Warmup epochs & {20} & 5 & 0 \\
		Weight decay & {0.05} & 0.05 & 1e-8 \\
		Label smoothing~\cite{label-smooth} & {0.1} & {0.1} & 0.1 \\
		Stochastic path~\cite{huang2016deep} & \ding{55} & \ding{55} & \ding{55} \\
		RandAugment~\cite{randaugment} & {$\checkmark$} & {$\checkmark$} & $\checkmark$ \\
		Mixup~\cite{mixup} & \ding{55} & \ding{55} & \ding{55} \\
		Cutmix~\cite{cutmix} & \ding{55} & \ding{55} & \ding{55} \\
		Random erasing~\cite{erasing} & {0.25} & {0.25} & \ding{55} \\
		EMA & {0.9999} & \ding{55} & 0.9999 \\
		\bottomrule[1pt]
	\end{tabular}
	\vspace{-0.em}
\end{table}

\paragraph{Complexity Analysis.}
For the standard convolutional layer, the number of parameters is $C_{out}\cdot C_{in}\cdot K\cdot K$ and the number of FLOPs are $H'\cdot W'\cdot C_{out}\cdot C_{in}\cdot K\cdot K$. The dynamic convolution consists of coefficient generation module, dynamic weight fusion and convolution process.
The coefficient generation module with $C_{in}$ hidden dimensions requires $C_{in}^2+C_{in}M$ parameters and $C_{in}^2+C_{in}M$ FLOPs. 
The dynamic weight fusion is parameter-free and has $M\cdot C_{out}\cdot C_{in}\cdot K\cdot K$ FLOPs. 
Thus, the total numbers of parameters and FLOPs of dynamic convolution are $C_{in}^2+C_{in}M + M\cdot C_{out}\cdot C_{in}\cdot K\cdot K$ and $C_{in}^2+C_{in}M + M\cdot C_{out}\cdot C_{in}\cdot K\cdot K+H'\cdot W'\cdot C_{out}\cdot C_{in}\cdot K\cdot K$ respectively. The parameter ratio of dynamic convolution over standard convolution is
\begin{equation}
	\begin{aligned}		
		R_{param} &= \frac{C_{in}^2+C_{in}M + M C_{out} C_{in} K^2}{C_{out} C_{in} K K} \\
		&= \frac{C_{in}}{C_{out} K^2} + \frac{M}{C_{out} K^2} + M\\
		&\approx \frac{1}{K^2} + M. \quad (M\ll C_{out} K^2,~C_{in}\approx C_{out})
	\end{aligned}
\end{equation}
The FLOPs ratio is
\begin{equation}
	\begin{aligned}
		R_{flops} &= \frac{C_{in}^2+C_{in}M + M C_{out} C_{in} K^2+H' W' C_{out} C_{in} K^2}{H' W' C_{out} C_{in} K^2} \\
		&= \frac{C_{in}}{H' W' C_{out} K^2} + \frac{M}{H' W' C_{out} K^2} \\
		& + \frac{M}{H' W'} + 1\\
		&\approx 1.  \quad (1<M\ll H'W',~C_{in}\approx C_{out})
	\end{aligned}
\end{equation}
Thus, compared to the standard convolution, the dynamic convolution has about $M\times$ parameters with negligible extra FLOPs. 

\subsection{Extending ParameterNet to Language Domain}
Sparse-activated Mixture-of-Experts (MoE) models~\cite{shazeer2017outrageously}, initially introduced in the NLP domain, allow for a substantial increase in the number of parameters while maintaining the computational load per token or sample unchanged. Numerous subsequent studies~\cite{zhou2022mixture,fedus2022switch,roller2021hash} have delved into exploring efficient routing mechanisms and have demonstrated the effectiveness of MoE in various large language models (LLMs) such as T5~\cite{t5}, NLLB~\cite{koishekenov2022memory}, LLaMA~\cite{llama} and Palm~\cite{palm}. In this context, our emphasis is primarily on low-FLOPs language models to validate the proposed hypothesis that incorporating more parameters can enhance the benefits of large-scale pretraining for low-FLOPs models, \ie, we proportionally reduce and construct a scaled-down version, LLaMA-1B.

Much like MoE, our approach involves taking a token representation, denoted as $x$, and subsequently routing it to the top-$k$ determined experts from a set of $N$. The router module generates logits represented as $h(x) = softmax(router(x))$, creating a normalized distribution through a softmax function over the available $N$ experts at that particular layer. The top-$k$ experts, where we consistently set $k=1$ in our experiments to maintain similar FLOPs to the original counterparts, are then selected for routing the token $x$. The training loss on expert capacity (the number of tokens each expert computes) follows the setting in Switch Transformer~\cite{fedus2022switch}.

\section{Experiment}
In this section, we conduct experiments to verify the effectiveness of the proposed ParameterNet scheme on visual pretrianing and extend it to language domain.

\begin{table*}[htb]
	\centering
	\small
	\renewcommand\arraystretch{1.1}
	\caption{ParameterNet results on ImageNet-1K val set by pretraining on ImageNet-1K and ImageNet-22K respectively.}
	\label{tab:ghostnet}
	\setlength{\tabcolsep}{10pt}
	\begin{tabular}{l|c|ccc}
		\toprule[1.5pt]
		Model & Pretrain data  & Parameters & FLOPs & Top-1  \\
		\midrule
		GhostNet-300M~\cite{ghostnet} & ImageNet-1K &  8.6M & 289M & 76.3 \\
		GhostNet-300M~\cite{ghostnet} & ImageNet-22K &  8.6M & 289M & 75.7 (\textbf{\color{red}-0.6}) \\
		ParameterNet-300M & ImageNet-1K &  15.7M & 298M & 77.2 \\
		ParameterNet-300M & ImageNet-22K &  15.7M & 298M & 79.1 (\textbf{\color{green}+1.9}) \\
		\midrule
		GhostNet-600M~\cite{ghostnet} & ImageNet-1K &  19.8M & 579M & 78.6 \\
		GhostNet-600M~\cite{ghostnet} & ImageNet-22K &  19.8M & 579M & 78.3 (\textbf{\color{red}-0.3}) \\
		ParameterNet-600M & ImageNet-1K &  34.5M & 599M & 79.0 \\
		ParameterNet-600M & ImageNet-22K &  34.5M & 599M & 81.6 (\textbf{\color{green}+2.6}) \\
		\bottomrule[1pt]
	\end{tabular}	
\end{table*}

\begin{table*}[htb]
	\centering
	\small
	\renewcommand\arraystretch{1.1}
	\caption{Comparison of ParameterNet and other SOTA models on ImageNet-1K val set. All the models are pretrained on large-scale visual datasets such as ImageNet-22K, JFT-300M and IG-1B-Targeted.}
	\label{tab:sota}	
	\setlength{\tabcolsep}{10pt}{
		\begin{tabular}{l|c|ccc}
			\toprule[1.5pt]
			Model & Pretrain data  & Parameters & FLOPs & Top-1  \\
			\midrule
			EfficientNet-B0~\cite{noisy-student} & JFT-300M & 5.3M & 390M & 78.1 \\
			Swin-300M~\cite{swin} & ImageNet-22K & 1.9M & 312M & 68.6 \\
			GhostNet-300M~\cite{ghostnet} & ImageNet-22K &  8.6M & 289M & 75.7 \\
			ParameterNet-300M & ImageNet-22K &  15.7M & 298M & \textbf{79.1} \\
			\midrule
			ResNet101~\cite{sun2017revisiting} & JFT-300M & 44.5M & 7.8G & 79.2 \\
			ResNet50 (Billion-scale)~\cite{yalniz2019billion} & IG-1B-Targeted & 25.6M & 4.1G & 81.2 \\
			ResNet50 (BiT)~\cite{bit} & ImageNet-22K & 25.6M & 12.0G & 80.0 \\
			EfficientNetV2-B0~\cite{efficientnetv2} & ImageNet-22K &  7.1M & 0.72G & 77.6 \\
			EfficientNetV2-B1~\cite{efficientnetv2} & ImageNet-22K &  8.1M & 1.2G & 79.0 \\
			Swin-T~\cite{swin} & ImageNet-22K & 28M & 4.5G & 80.9 \\
			GhostNet-600M~\cite{ghostnet} & ImageNet-22K &  19.8M & 579M & 78.3 \\
			ParameterNet-600M & ImageNet-22K &  34.5M & 599M & \textbf{81.6} \\
			\bottomrule[1pt]
		\end{tabular}
	}	
\end{table*}

\subsection{Experimental Settings}
\paragraph{Datasets.}
We adopt the widely-used ImageNet-22K for large-scale pretraing and ImageNet-1K as the normal training data for comparison. ImageNet-22K~\cite{imagenet} is a large-scale image dataset with 14,197,122 images belonging to 21841 categories. ImageNet-1K~\cite{imagenet} is a subset of ImageNet-22K with 1,000 object classes. It contains 1,281,167 training images, and 50,000 validation images.

\begin{figure*}[htb]
	\centering	
	\setlength{\tabcolsep}{4pt}{
		\begin{tabular}{cc}
			\makecell*[c]{\includegraphics[width=0.5\linewidth]{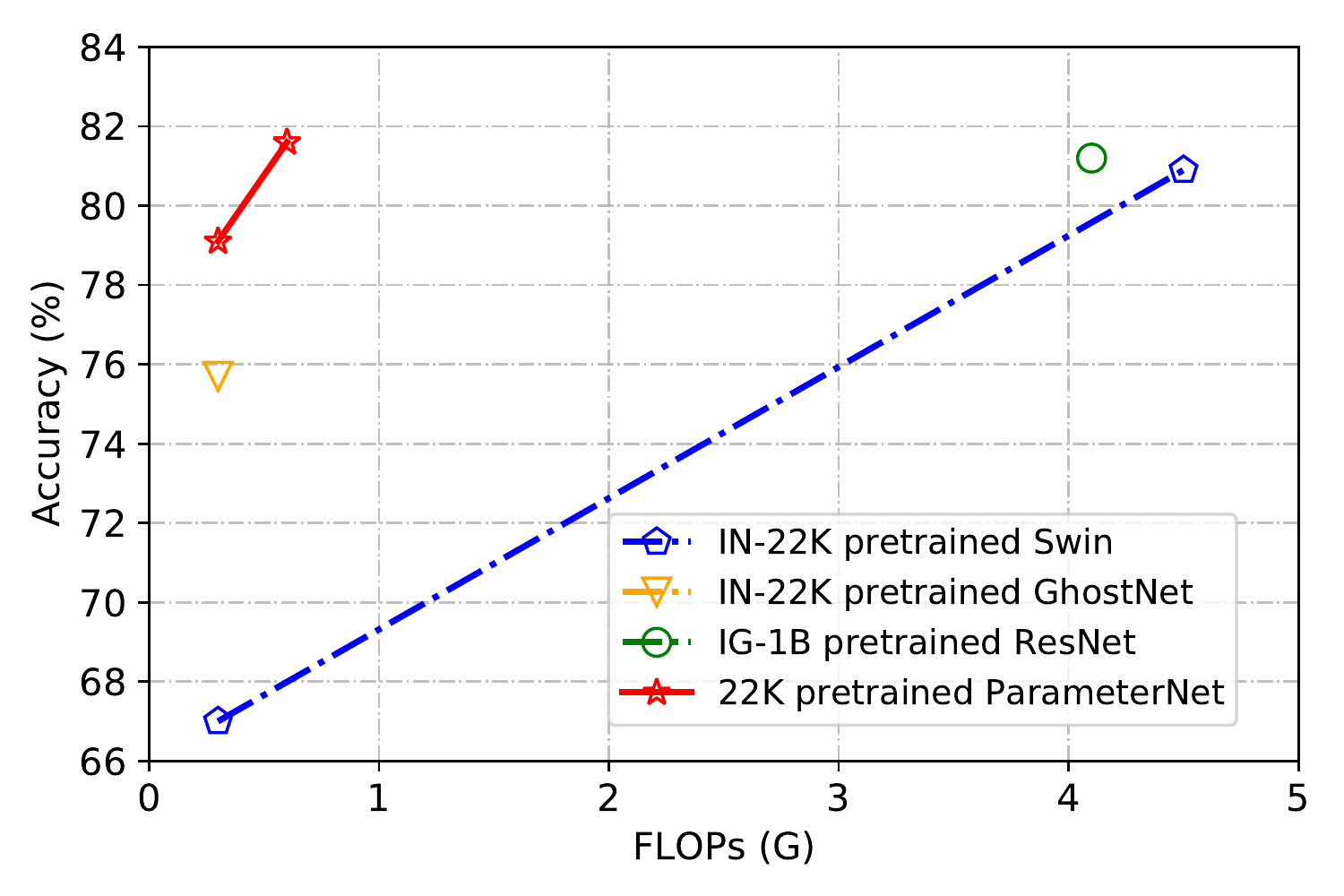}}  & \makecell*[c]{\includegraphics[width=0.5\linewidth]{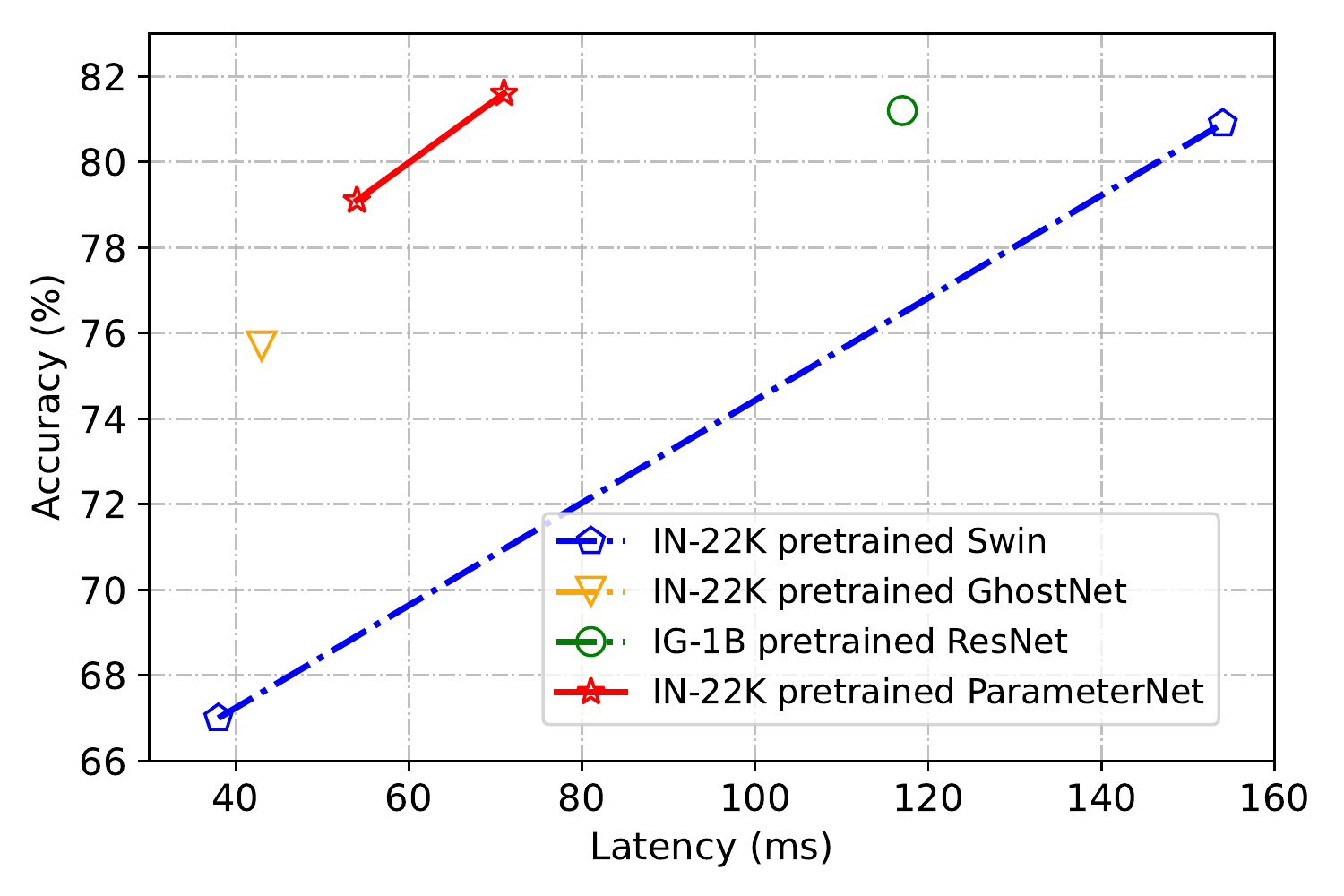}} 
			\\
			\small (a) Acc v.s. FLOPs & \small (b) Acc v.s. Latency
		\end{tabular}
	}
	\caption{Performance comparison of the representative visual backbone networks with ImageNet-22K pretraining.}
	\label{fig:latency}
	\vspace{-1em}
\end{figure*}

\paragraph{Training on ImageNet-1K.}
Following the common setting~\cite{deit,swin,convnext}, we train the models for 300 epochs
with 20 warm-up epochs using AdamW~\cite{adamw} optimizer. We use a batch size of 1024. The base learning rate is set as $0.001$ and decays with the cosine schedule. The data augmentation strategy includes RandAugment~\cite{randaugment} and random erasing~\cite{erasing}. Weight decay and label smoothing are adopted for regularization. More details can be found in Table~\ref{table-hyper}.

\paragraph{Pretraining on ImageNet-22K.}
The models are pretraining on ImageNet-22K for 90 epochs with 5 warm-up epochs. The batch size is $4096$ and The base learning rate is set as $0.004$. Other settings basically follow those on ImageNet-1K as shown in Table~\ref{table-hyper}.

\paragraph{Finetuning on ImageNet-1K.}
We finetune the pretrained models on ImageNet-1K for 30 epochs without warm-up epochs. The batch size is $512$ and The base learning rate is set as $0.0005$. The weight decay is set as 1e-8 and random erasing~\cite{erasing} is switched off for better fitting on ImageNet-1K. Other settings basically follow those on ImageNet-1K as shown in Table~\ref{table-hyper}.

\subsection{Main Results on Vision Domain}
We build the baseline GhostNet with different FLOPs (\ie, $\sim$300M and $\sim$600M) by tuning the width and depth. Our ParameterNet is constructed by replacing the conventional convolutional layer with dynamic convolution. The number of experts is set as 4 by default. The details of the network architectures are available in the appendix. The results are shown in Table~\ref{tab:ghostnet}. Training only on ImageNet-1K, ParameterNet outperforms the original GhostNet by 0.4-xx accuracy. For GhostNet, pretraining on ImageNet-22K does not help to the performance. ImageNet-22K pretrained ParameterNet can achieve more than 2\% improvement over ImageNet-1K. This indicates that our ParameterNet with more parameters yet similar FLOPs can benefit from the large-scale visual pretraining.

\paragraph{Comparison with SOTA.}
We compare ParameterNet with other representative models pretrained on ImageNet-22K or larger datasets such as JFT-300M~\cite{sun2017revisiting} and IG-1B-Targeted~\cite{yalniz2019billion}. From the results in Table~\ref{tab:sota}, we can see that our ParameterNet with fewer FLOPs outperforms other models pretrained on large-scale datasets. For example, ParameterNet-600M achieves 81.6\% top-1 accuracy whose \#FLOPs is about 7$\times$ lower than that of ResNet50 or Swin-T.

\paragraph{Inference speed.}
We evaluate the inference speed of ParameterNet and other representative models for comparison. We run models using ONNX toolkit on Intel Xeon Platinum 8378C CPU with single-thread mode. As shown in Figure~\ref{fig:latency}, our ParameterNet outperforms the widely-used ResNet and Swin Transformer for much better accuracy-latency trade-off.

\begin{table}[htb]
	\centering
	\small
	\renewcommand\arraystretch{1.1}
	\caption{ImageNet-1K val set results \wrt~\#Expert. The base network architecture is GhostNet-300M.}
	\label{tab:expert}
	\begin{tabular}{c|c|ccc}
		\toprule[1pt]
		\#Expert & Pretrain data  & Parameters & FLOPs & Top-1  \\
		\midrule
		1 & ImageNet-1K &  8.6M & 289M & 76.3 \\
		1 & ImageNet-22K &  8.6M & 289M & 75.7 \\
		\midrule
		2 & ImageNet-1K &  11.0M & 293M & 76.9 \\
		2 & ImageNet-22K &  11.0M & 293M & 77.7 \\
		\midrule
		4 & ImageNet-1K &  15.7M & 298M & 77.2 \\
		4 & ImageNet-22K &  15.7M & 298M & 79.1 \\
		\midrule
		8 & ImageNet-1K &  25.2M & 308M & 77.7 \\
		8 & ImageNet-22K &  25.2M & 308M & 79.4 \\
		\bottomrule[1pt]
	\end{tabular}	
\end{table}

\subsection{Ablation Study}
\paragraph{The number of dynamic experts.}
The number of dynamic experts is an important hyperparameter of dynamic convolution, which directly controls the parameters and FLOPs. As shown in Table~\ref{tab:expert}, more experts will largely increase the number of parameters and slightly influence FLOPs. The performance of more experts improves over fewer experts. We use 4 experts by default for efficiency trade-off.

\begin{table}[htb]
	\centering
	\small
	\renewcommand\arraystretch{1.1}
	\caption{Results on ImageNet-1K val set. The base network architecture is Swin-300M.}
	\label{tab:swin}
	\setlength{\tabcolsep}{4pt}
	\begin{tabular}{l|c|ccc}
		\toprule[1.5pt]
		& Pretrain & Parameters & FLOPs & Top-1  \\
		\midrule
		Original & ImageNet-1K &  1.9M & 312M & 69.2 \\
		Original & ImageNet-22K &  1.9M & 312M & 67.0 (\textbf{\color{red}-2.2}) \\
		Ours & ImageNet-1K &  6.8M & 323M & 72.3 \\
		Ours & ImageNet-22K &  6.8M & 323M & 74.5 (\textbf{\color{green}+2.2}) \\
		\bottomrule[1pt]
	\end{tabular}	
\end{table}

\begin{table*}[htb]
	\centering
	\small
	\renewcommand\arraystretch{1.1}
	\setlength{\tabcolsep}{8pt}
	\caption{Comparison of different approaches to construct ParameterNet on ImageNet-1K val set. The base network architecture is GhostNet-300M.}
	\label{tab:rep}
	\begin{tabular}{c|c|cccc}
		\toprule[1.5pt]
		Method & Pretrain data  & Train parameters  & Inference parameters & FLOPs & Top-1  \\
		\midrule
		RepConv & ImageNet-1K &  15.7M & 8.6M & 289M & 76.8 \\
		RepConv & ImageNet-22K &  15.7M & 8.6M & 289M & 76.9 \\
		DynamicConv & ImageNet-1K & 15.7M & 15.7M & 298M & 77.2 \\
		DynamicConv & ImageNet-22K & 15.7M & 15.7M & 298M & 79.1 \\
		\bottomrule[1pt]
	\end{tabular}	
\end{table*}

\paragraph{Dynamic convolution \emph{vs.} re-parameterized convolution.}
As we discussed before, there are various approaches to construct ParameterNet, such as dynamic convolution~\cite{dynamic-conv} and re-parameterized convolution~\cite{repvgg}. We compare these two approaches where the dynamic convolution has 4 experts and the re-parameterized convolution has 3 more paralleled branches based on the original convolution. From the results in Table~\ref{tab:rep}, although the re-parameterized convolution increase the training parameters, its parameters and FLOPs are unchanged for inference, that is, the model capacity is not increased and the ImageNet-22K pretrained performance does not improve.

\paragraph{ParameterNet for other network architectures.}
In addition to CNN, we extend ParameterNet to the transformer architecture (\ie, Swin Transformer). To construct a smaller version, we set the token dimension of Swin-T to 24 to obtain Swin-300M with about 300M FLOPs. From the results in Table~\ref{tab:swin}, the original Swin-300M has a significant accuracy drop when pretraining on ImageNet-22K. Our strategy can achieve +2.2\% performance gain from ImageNet-22K pretraining.

\begin{table*}[htb]
	\centering
	\small
	\renewcommand\arraystretch{1.1}
	\caption{\small{ParameterNet (sparse-activated MoE) on LLaMA-1B. Given the SwiGLU activation function in LLaMA, there are three linear projections in the FFN module. We added parameters at each linear projection, and we present the corresponding zero-shot results, except for SST-2 (where we fine-tuned the classifier).} The best results are in \textbf{bold} and the second best are \underline{underlined}.}
	\label{tab:nlp}	
	\setlength{\tabcolsep}{7pt}{
		\begin{tabular}{l|c|cc|c|cccc|c}
			\toprule[1.5pt]
			Model  & \#Expert & Parameters & FLOPs & Training loss & ARC (easy) & BoolQ & SST-2 & HellaSwag & Avg \\
			\midrule
			LLaMA-1B & - & 0.94B & 919G & 1.86 & 47.99 & 57.31 & 88.53 & 38.92 & 58.19 \\
			\midrule
			MoE on gate & 4 & 1.54B & 919G & 1.75 & 48.81 & 57.86 & 88.88 & 42.09 & 59.41 \\
			MoE on gate & 8 & 2.35B & 919G & 1.64 & 49.04 & 57.70 & 89.56 & \underline{43.78} & 60.02 \\
			\midrule
			MoE on up proj & 4 & 1.54B & 919G & 1.72 & 49.09 & 58.04 & 89.67 & 42.35 & 59.79 \\
			MoE on up proj & 8 & 2.35B & 919G & 1.61 & \textbf{49.58} & \underline{58.39} & \underline{90.05} & \textbf{44.21} & \textbf{60.56} \\
			\midrule
			MoE on down proj & 4 & 1.54B & 919G & 1.70 & \underline{49.36} & 57.99 & 89.13 & 41.95 & 59.61 \\
			MoE on down proj & 8 & 2.35B & 919G & 1.62 & 49.22 & \textbf{58.58} & \textbf{90.34} & 43.60 & \underline{60.44} \\
			\bottomrule[1pt]
		\end{tabular}
	}	
\end{table*}

\subsection{Extensive Experiment on Language Domain}

\paragraph{Datasets.}
Our training dataset is a mixture of several sources, including C4~\cite{c4}, Wikipedia~\cite{llama}, and ArXiv~\cite{arxiv}. The data are all publicly available, and we directly mix them without any quality filtering. Overall, the training dataset contains roughly 90B tokens after tokenization. Each token is used only once during training. The learning rate is set to 0.0003 with a batch size of 4M (input token length is 2048). We use the AdamW optimizer with a cosine learning rate schedule, ensuring that the final learning rate is equal to 10\% of the maximal learning rate.

\paragraph{Network architecture.}
We build a baseline LLaMA-1B by proportionally reduce the dimension and the number of layers based on original LLaMA~\cite{llama}, as shown in Table~\ref{tab:llama}. Specifically, the hidden size, intermediate size, number of head, and the number of layer are 2048, 8191, 16 and 12, respectively. The tokenizer is the same with LLaMA.

\begin{table}[htb]
	\centering
	\small
	\renewcommand\arraystretch{1.1}
	\caption{Network architecture of LLaMA-1B baseline. In the following experiments, we equipe the fully-connected layer in FFN with MoE to verify the proposed method.}
	\label{tab:llama}
	\setlength{\tabcolsep}{3pt}
	\begin{tabular}{l|ccccc}
		\toprule[1.5pt]
		Model & Dimension & Heads & Layers & Parameters & FLOPs \\
		\midrule
		LLaMA-1B & 2048 & 16 & 12 & 0.94B & 919G \\
		\bottomrule[1pt]
	\end{tabular}
	\vspace{-1em}	
\end{table}

\paragraph{Results and analysis.}
Following previous work~\cite{brown2020language}, we present the corresponding training loss and zero-shot results on several common-sense reasoning tasks, where the model ranks the proposed answers. FLOPs are calculated with the output response length set to 1. The router module is implemented with a linear layer, with the input channel being the hidden size and the output channel equal to the number of experts. As shown in Table~\ref{tab:nlp}, we observe that more experts bring additional parameters to the baseline model, leading to a noticeable improvement in downstream performance. For example, LLaMA-1B with 8 experts on up projection layers obtains a 2.37\% accuracy gain on average. Moreover, the increased parameters help reduce the training loss, indicating enhanced understanding of the input data by incorporating ParameterNet into the language model. Additionally, experimental results suggest that the three linear projections in LLaMA's FFN have similar effects.

\section{Conclusion}
In this paper, we propose a design principle (\ie, ParameterNet) for large-scale visual pretraining by adding more parameters while maintaining low FLOPs.
ParameterNet is a general scheme and has various approaches to implement such as dynamic convolution and re-parameterized convolution. We use the dynamic convolution in practice to construct the ParameterNet models. ParameterNet can overcome the \emph{low FLOPs pitfall} and much benefit from large-scale visual pretraining.
The experiments on ImageNet-22K large-scale dataset have demonstrated the effectiveness of the proposed ParameterNet. We also verify the generalization of our method on language domain.
We hope our work can motivate and inspire the future research on vision, multimodality and large language models.

{
    \small
    \bibliographystyle{ieeenat_fullname}
    \bibliography{ref}
}


\end{document}

%% file: preamble.tex
\usepackage[dvipsnames]{xcolor}
